\documentclass[11pt]{article}

\usepackage[final]{latex/acl}

\usepackage{times}
\usepackage{latexsym}

\usepackage{subcaption}

\usepackage[T1]{fontenc}

\usepackage[utf8]{inputenc}

\usepackage{microtype}

\usepackage{inconsolata}

\usepackage{graphicx}

\usepackage{hyperref}
\usepackage{color}
\usepackage{makecell}
\usepackage{amsmath}
\usepackage{verbatim}
\usepackage{caption}
\usepackage{subcaption} 

\usepackage{booktabs,siunitx,multirow}
\newcolumntype{N}{S[table-format=1.2, table-number-alignment=center, table-text-alignment=center]}

\title{SegNSP: Revisiting Next Sentence Prediction for\\
Linear Text Segmentation}

\author{
  {\bf José Isidro}$^{1,2\dagger}$, {\bf Filipe Cunha}$^{1,2\dagger}$, {\bf Purificação Silvano}$^{1,2\dagger}$, \\
  {\bf Alípio Jorge}$^{1,2\dagger}$, {\bf Nuno Guimarães}$^{1,2\dagger}$, {\bf Sérgio Nunes}$^{1,2\dagger}$, \and {\bf Ricardo Campos}$^{2,3\dagger}$ \\
  \textnormal{$^{1}$University of Porto, Porto, Portugal} \\
  \textnormal{$^{2}$INESC TEC, Porto, Portugal} \\
  \texttt{\{jose.m.isidro, luis.f.cunha, purificacao.silvano, alipio.jorge,} \\
  \texttt{nuno.r.guimaraes, sergio.nunes\}@inesctec.pt} \\
  \textnormal{$^{3}$University of Beira Interior, Covilhã, Portugal} \\
  \texttt{ricardo.campos@ubi.pt}
}

\begin{document}
\maketitle
\begin{abstract}
Linear text segmentation is a long-standing problem in natural language processing (NLP), focused on dividing continuous text into coherent and semantically meaningful units. Despite its importance, the task remains challenging due to the complexity of defining topic boundaries, the variability in discourse structure, and the need to balance local coherence with global context. These difficulties hinder downstream applications such as summarization, information retrieval, and question answering. In this work, we introduce SegNSP, framing linear text segmentation as a next sentence prediction (NSP) task. Although NSP has largely been abandoned in modern pre-training, its explicit modeling of sentence-to-sentence continuity makes it a natural fit for detecting topic boundaries. We propose a label-agnostic NSP approach, which predicts whether the next sentence continues the current topic without requiring explicit topic labels, and enhance it with a segmentation-aware loss combined with harder negative sampling to better capture discourse continuity. Unlike recent proposals that leverage NSP alongside auxiliary topic classification, our approach avoids task-specific supervision. We evaluate our model against established baselines on two datasets, CitiLink-Minutes, for which we establish the first segmentation benchmark, and WikiSection. On CitiLink-Minutes, SegNSP achieves a B-$F_1$ of 0.79, closely aligning with human-annotated topic transitions, while on WikiSection it attains a  B-F$_1$ of 0.65, outperforming the strongest reproducible baseline, TopSeg, by 0.17 absolute points. These results demonstrate competitive and robust performance, highlighting the effectiveness of modeling sentence-to-sentence continuity for improving segmentation quality and supporting downstream NLP applications.
\end{abstract}

\section{Introduction}
\label{sec:introduction}
Linear text segmentation (LTS) is a fundamental task in natural language processing (NLP), focused on dividing continuous text into contiguous, topically coherent segments in a linear, non-hierarchical sequence. By imposing structure on otherwise unsegmented text, LTS enables the effective processing and interpretation of large volumes of unstructured data. This structural decomposition is central to a variety of downstream NLP applications. In document summarization, segmentation supports the identification and aggregation of topic-consistent content units \cite{cho2022toward} contributing to more coherent and informative summaries. In information retrieval (IR) \cite{tiedemannmur2008simple}, and question answering \cite{oh2007semantic} it facilitates the construction of semantically coherent text chunks, which are critical for dense retrieval \cite{liu2025passage} and retrieval-augmented generation (RAG) pipelines \cite{duarte2024lumberchunker}.

A common formulation of LTS models a document as a linear sequence of topics and focuses on detecting boundaries between consecutive segments.
In this view, a document consists of consecutive sentences grouped into contiguous segments, each corresponding to a single topic. Segment boundaries indicate points where the topic changes.

\begin{figure}[h]
    \centering
    \includegraphics[width=0.9\linewidth]{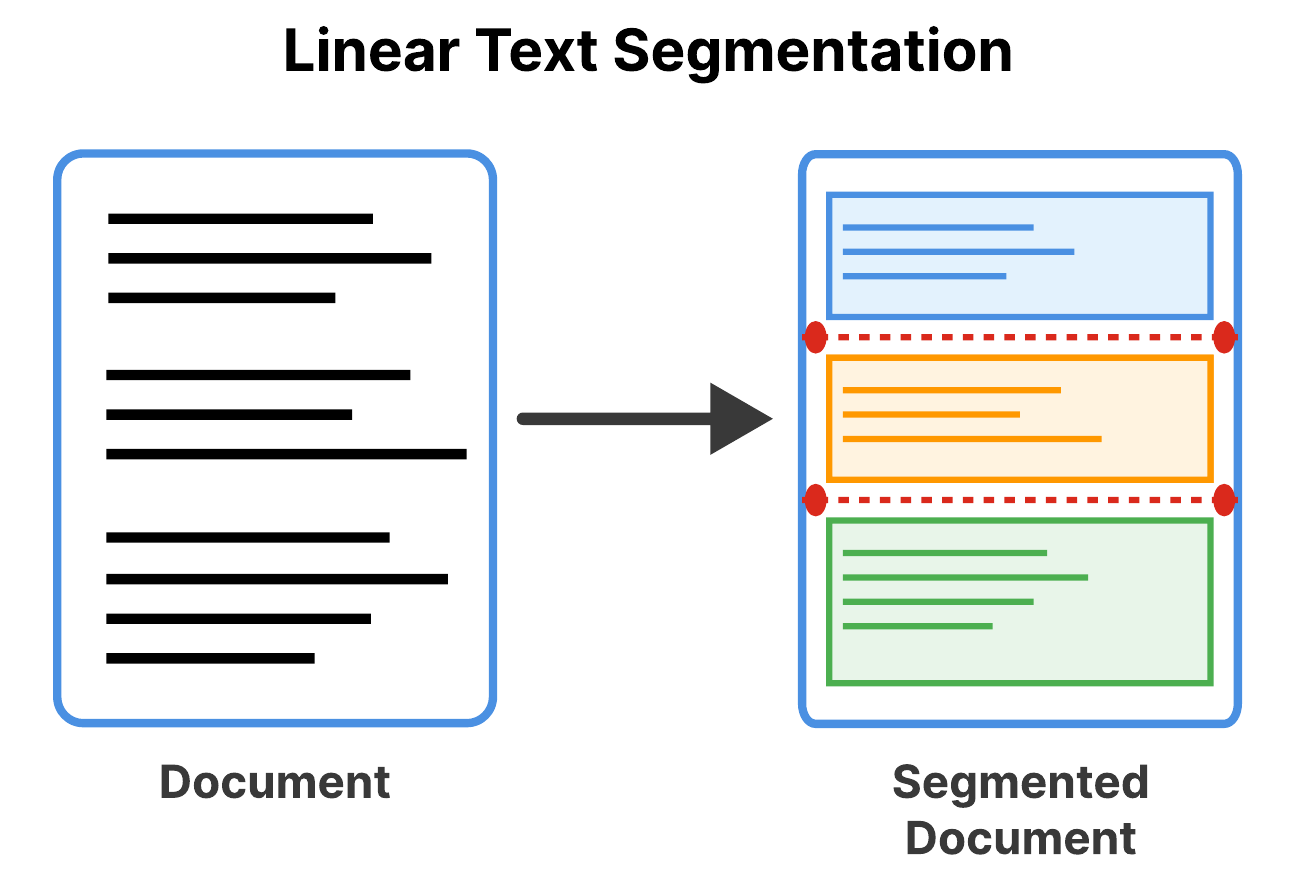}
    \caption{Overview of linear text segmentation: A document is first split into individual sentences, which are then grouped into contiguous segments corresponding to coherent topical units.}
    \label{fig:segments_diagram}
\end{figure}

Figure~\ref{fig:segments_diagram} provides a schematic view of this formulation, illustrating how an unsegmented document is partitioned into contiguous, topically coherent segments, with boundaries indicating shifts in topics.

Despite its importance, LTS remains a challenging task in practice. Segment boundaries are often gradual, subjective, and domain-dependent \cite{eisenstein2008bayesian,galley2003discourse}, which makes the task inherently ambiguous. These factors help explain why both heuristic and model-based approaches struggle to achieve robust and generalizable performance \cite{glavas2016unsupervised,riedl2012topictiling}, particularly on long, unstructured documents, such as legal texts \cite{aumiller2021legal,malik2022semantic} or meeting transcripts \cite{solbiati2021unsupervised}, where topic shifts are often not explicitly marked.

Recent advances in contextual language modeling have renewed interest in leveraging the continuity between consecutive sentences as a proxy for topical coherence \cite{iter2020pretraining}. However, many modern approaches focus primarily on token-level prediction \cite{liu2019roberta}, which often overlooks how sentences relate to one another across the text. This limitation suggests that objectives explicitly modeling sentence-to-sentence dependencies could provide a natural signal for detecting topic boundaries. In this work, we revisit the \emph{next sentence prediction} (NSP) objective as a foundation for LTS. We frame NSP as a sentence-pair classification problem: given consecutive sentences from a document, a language model fine-tuned with this objective predicts whether the second sentence follows the first or marks a topical boundary. By fine-tuning pretrained language models on domain-specific corpora, we adapt NSP to capture discourse continuity across domains and languages.

We evaluate our approach on WikiSection \cite{arnold2019sector}, a well-known dataset of Wikipedia articles with manually annotated section boundaries, and extend the evaluation to an underexplored administrative domain and language using CitiLink-Minutes \cite{citilink2025}, a Portuguese dataset of city council meeting minutes, providing the first systematic evaluation of text segmentation on this dataset. Our experiments show that NSP achieves competitive performance compared to established baselines, challenging assumptions about its obsolescence and highlighting its effectiveness as a lightweight approach for linear text segmentation. 

The main contributions of this work are as follows:

\begin{itemize}
    \item We revisit the classic Next Sentence Prediction (NSP) objective for linear text segmentation, demonstrating that, when adapted with a segmentation-aware formulation and domain-specific fine-tuning, it remains a competitive and lightweight approach.
    \item We conduct a comprehensive empirical evaluation of NSP-based segmentation on both established benchmarks and deliberative governance data, comparing it against strong lexical and neural baselines using standard segmentation metrics.
    \item We provide the first systematic evaluation of text segmentation in an underexplored administrative domain and non-English language, using CitiLink-Minutes, a Portuguese dataset of city council meeting minutes, thereby establishing a benchmark for future research.
    \item We show that NSP, often considered obsolete in modern pretraining, can effectively model sentence-to-sentence continuity, highlighting its potential for robust, cross-domain, and cross-lingual segmentation applications.
\end{itemize}

Additionally, we release the experimental code\footnote{\url{https://github.com/anonymous15135/revisiting-NSP-for-LTS}}, the fine-tuned NSP models\footnote{\url{https://huggingface.co/anonymous15135/models}}, and an interactive demo\footnote{\url{https://huggingface.co/spaces/anonymous15135/nsp-councilseg-demo}}, enabling further research on this topic. 

The remainder of this paper is organized as follows. Section~\ref{sec:related_work} reviews related work on linear text segmentation, with particular emphasis on sentence-continuity modeling and neural boundary detection approaches. Section~\ref{sec:our_approach} describes our NSP-based method for text segmentation. Section~\ref{sec:experimental_setup} details the experimental setup, and Section~\ref{sec:experiments_and_results} presents the results. Finally, Section~\ref{sec:conclusions} concludes the paper and outlines directions for future work.

\section{Related Work}
\label{sec:related_work}

Research on linear text segmentation has evolved from early unsupervised lexical-cohesion methods~\cite{choi2000advances,hearst1997texttiling} to modern neural and large language model (LLM)-based approaches, such as zero-shot prompting of GPT-4~\cite{wang2024problem}, and dynamic chunking~\cite{duarte2024lumberchunker}. Early unsupervised approaches used surface cues and windowed similarity measures (e.g., Hearst’s TextTiling \cite{hearst1997texttiling} and Choi’s C99~\cite{choi2000advances}). Probabilistic and topic modeling methods such as BayesSeg~\cite{eisenstein2008bayesian} and LDA-based approaches~\cite{blei2003latent} introduced generative models that explicitly represent topical structure.

More recently, embedding-based unsupervised methods, including Content Vector Segmentation~\cite{alemi2015} and GraphSeg~\cite{glavas2016unsupervised}, leveraged distributed representations to capture semantic coherence beyond surface cues. In parallel, supervised methods framed segmentation as a boundary prediction task~\cite{koshorek2018textseg}, and were subsequently enhanced by attention-based and hierarchical architectures that explicitly model sentence-level coherence~\cite{badjatiya2018attention,riedl2012topictiling}.

The introduction of Transformers~\cite{vaswani2017attention} enabled methods such as Cross-Segment BERT~\cite{lukasik2020cross}, and other Transformer-based classifiers~\cite{glavas2021training,yu2023improving} which combine local and global context to substantially improve boundary detection. Multi-task strategies, such as jointly training on segmentation and topic classification objectives, further enhanced robustness and generalization across domains~\cite{lee2023}.

Building on recent advances, LLMs have demonstrated strong zero and few-shot capabilities for topic segmentation through prompting and contextual reasoning~\cite{ghinassi2024survey}. For example, \citet{duarte2024lumberchunker} introduced LumberChunker, which uses LLMs to dynamically segment documents into semantically coherent chunks, significantly improving dense retrieval performance in RAG systems. Despite these advances, LLM-based segmentation still faces significant challenges, such as hallucinations, high computational costs, inconsistent segmentation on domain-specific corpora, and imprecise segment offsets. These limitations, highlighted by \citet{ghinassi2024survey}, have renewed interest in Transformer-based encoders and supervised approaches, which offer more controllable and efficient solutions for segmentation tasks. A particular area of resurgence concerns pre-training objectives that explicitly model discourse-level coherence. Among these, NSP, originally part of BERT’s pre-training but later dropped in models such as RoBERTa~\cite{liu2019roberta}, has recently attracted revived attention. In this context, \citet{nspbert} demonstrated that NSP can yield competitive performance across downstream tasks and be effectively adapted to domain-specific settings, achieving strong results in modeling fine-grained discourse coherence. Related objectives, such as sentence order prediction, have likewise been explored to improve document-level representations~\cite{cui2020bert,lan2019albert}. Overall, these renewed directions highlight that sentence-continuity objectives remain a viable mechanism for modeling discourse structure. However, existing work has primarily focused on general-purpose benchmarks and document understanding tasks, leaving their applicability to linear text segmentation, particularly for long, domain-specific documents, largely underexplored. Motivated by this gap in the literature, we revisit NSP as a lightweight and controllable objective specifically tailored to the segmentation task.

\section{Our Approach}
\label{sec:our_approach}
In this work, we adapt next sentence prediction (NSP) objective for document segmentation, motivated by its demonstrated ability to model continuity between adjacent sentences. We hypothesize that, with sufficient domain-specific fine-tuning, NSP can distinguish natural continuations within a topical unit from transitions across topics. This hypothesis is motivated by recent work by \citet{lee2023}, who combined NSP with same-topic prediction and topic classification in a multi-task framework, achieving state-of-the-art results on WikiSection. Despite these results, multi-task approaches, such as those of \citet{lee2023}, introduce additional architectural complexity and require the joint optimization of auxiliary classification objectives, which can limit portability to domains without a fixed or consistent topic taxonomy. In contrast, we propose a label-agnostic NSP-based model that treats segmentation as a primary coherence task. By focusing on continuity between consecutive sentences, our approach provides a more streamlined alternative that remains effective without the computational overhead of auxiliary tasks, such as topic classification.

Building upon this, a document $D$ is represented as an ordered sequence of sentences (or utterances),
\[
D = (s_1, s_2, \dots, s_n)
\]
where each $s_i$ denotes a single sentence. The objective of LTS consists of partitioning $D$ into $m$ contiguous, topically coherent segments
\[
G = (seg_1, \dots, seg_m)
\]
where each segment
\[
seg_j = (s_{a_j}, \dots, s_{b_j)}
\]
is a subsequence of consecutive sentences discussing a common topic.  Segment boundaries correspond to the transition points between adjacent segments, that is, between the last sentence of $seg_j$ and the first sentence of $seg_{j+1}$.

Given this formulation, we cast linear text segmentation as a sentence-pair classification problem grounded in the Next Sentence Prediction (NSP) objective. For each pair of consecutive sentences $(s_i, s_{i+1})$ in a document $D$, a language model fine-tuned for NSP produces a probabilistic prediction
\[
f_\theta : (s_i, s_{i+1}) \mapsto \hat{y}_i,
\]
where $\hat{y}_i \in [0,1]$ represents the likelihood that $s_{i+1}$ continues the current topical segment. This framing models LTS to a boundary detection problem based on local discourse coherence, allowing the global segment structure to be reconstructed from sentence-level continuity signals.

We instantiate the function $f_\theta$ with a pretrained BERT model fine-tuned with the NSP objective. Each sentence pair $(s_i, s_{i+1})$ is encoded by concatenating the two sentences $s_i$ and $s_{i+1}$ and passing them through the model, producing a joint representation of the input sequence: 
\begin{equation*}
\mathbf{h}_{[\texttt{CLS}]} =
\mathrm{BERT}([\texttt{CLS}] \; s_i \; [\texttt{SEP}] \; s_{i+1} \; [\texttt{SEP}])
\end{equation*}

\begin{figure*}[t]
    \centering
    \includegraphics[width=1\textwidth]{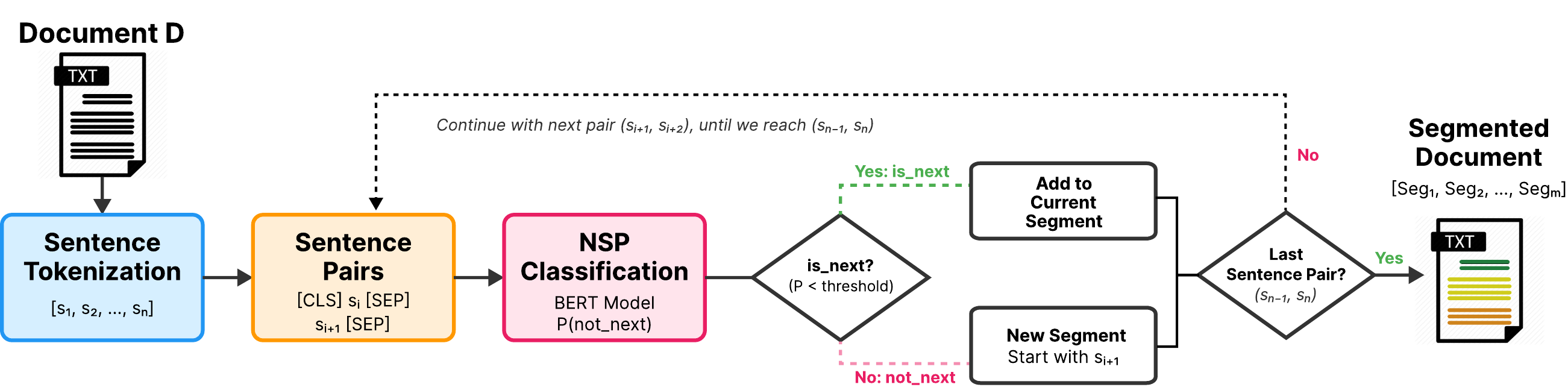}
    \caption{Overview of the NSP-based segmentation pipeline.}
    \label{fig:nsp_diagram}
\end{figure*}

This representation is then fed into a linear classifier, and the resulting logits are normalized with a softmax function to obtain NSP class probabilities:
\begin{equation*}
P(y \mid s_i, s_{i+1}) =
\mathrm{softmax}\big(W \mathbf{h}_{[\texttt{CLS}]} + b\big),
\end{equation*}

where $y \in \{\texttt{is\_next}, \texttt{not\_next}\}$. During inference, a boundary is predicted at position $i$ whenever the probability of the \texttt{not\_next} class exceeds a decision threshold  $\tau$ tuned on validation data, i.e.,
\[
P(y = \texttt{not\_next} \mid s_i, s_{i+1}) > \tau
\]
Contiguous spans between predicted boundaries then form coherent segments, reconstructing the document's global topical structure. Figure~\ref{fig:nsp_diagram} illustrates the end-to-end pipeline.

While the original BERT NSP formulation was designed for general pre-training and often relied on random sentence sampling, adapting it for linear text segmentation requires specific modifications to handle the sparsity and subtlety of topical boundaries. To this end, our approach introduces \textbf{domain-balanced ratios}, \textbf{harder negative sampling} and a \textbf{segmentation-aware loss}, effectively redefining NSP as a boundary-sensitive coherence classifier rather than a generic pre-training task:

\begin{itemize}
    \item \textbf{Domain-Balanced Ratios}: In natural discourse, segment boundaries are sparse events. A model trained on raw consecutive pairs would mostly see "same-segment" examples, leading to conservative predictions and causing the model to frequently miss true topic transitions. By enforcing a fixed 70/30 distribution of intra-segment (positive) and inter-segment (boundary) pairs, we ensure the model receives sufficient exposure to transition signals, improving learning of topical shifts.
    \item \textbf{Harder Negative Sampling}: Although the 30\% inter-segment pairs provide the basic boundary signal, they often represent obvious shifts between distant topics. To improve robustness against "false boundary cues" and over-segmentation, we introduce "hard negatives" - up to 10 randomly mismatched sentence pairs from within the same document. Unlike standard inter-segment pairs, which are adjacent, these "hard" negatives force the model to focus on local semantic continuity rather than superficial document-level cues.
    \item \textbf{Segmentation-Aware Loss.}  
    To further emphasize difficult boundary decisions, we employ a segmentation-aware loss function:
    \[
    L_{\mathrm{seg}} = L_{\mathrm{focal}} + \lambda_{1} L_{\mathrm{conf}} + \lambda_{2} L_{\mathrm{bound}},
    \]
    where $L_{\mathrm{focal}}$ addresses class imbalance, $L_{\mathrm{conf}}$ penalizes overconfident incorrect predictions, and $L_{\mathrm{bound}}$ assigns higher weights to errors near true boundaries. This formulation encourages the model to prioritize the correct classification of challenging transition points, which is critical for boundary-based evaluation metrics.
\end{itemize}

During training, the model learns to discriminate between sentence pairs that continue the current topical segment and those that span a boundary, under the adapted NSP formulation and segmentation-aware loss. At inference time, boundary probabilities $\hat{y}_i$ are computed for each consecutive sentence pair $(s_i, s_{i+1})$ and compared against a decision threshold $\tau$ to recover the document’s segmentation. In the following section, we describe our experimental setup and evaluation protocols to assess the model's performance and generalizability across domains and languages.

\section{Experimental Setup}
\label{sec:experimental_setup}

\paragraph{Datasets}
We evaluate our approach on two datasets that represent complementary evaluation scenarios. WikiSection \cite{arnold2019sector} is a widely used benchmark for topic segmentation, derived from Wikipedia articles with explicit section boundaries. These boundaries provide a reliable approximation of topic shifts, making the dataset particularly suitable for segmentation evaluation. WikiSection exhibits relatively regular discourse, where topics are developed coherently and hierarchically. Its broad topical coverage (two categories: disease and city) and extensive use in prior work enable direct comparison with existing approaches, supporting reproducibility and fair benchmarking. In our experiments, we focus on the \textit{en\_city} subset of WikiSection, the largest of the two available categories, comprising 19,539 documents and 133,642 annotated segments.

In contrast, CitiLink-Minutes \cite{citilink2025} comprises 120 Portuguese city council meeting minutes collected from six municipalities between 2021 and 2024. Unlike WikiSection, topic boundaries in this corpus are often implicit, embedded in procedural language, enumerations, and shifts between administrative and deliberative content. This results in a higher degree of structural variability and noise, making the dataset a challenging testbed for segmentation models beyond well-structured text explicitly marked by headings. Each municipality contributes with 20 minutes, manually annotated with 2,848 subjects of discussion. To adapt the dataset for linear text segmentation, we reconstruct document segments from agenda item markers, which serve as implicit topical delimiters, treating each agenda heading and its associated text as a contiguous segment. This approach preserves the deliberative structure while aligning annotations with segmentation units, allowing CitiLink-Minutes to be used as a realistic benchmark for evaluating segmentation in institutional meeting records.

Together, these datasets provide a complementary evaluation framework. WikiSection enables controlled benchmarking against existing methods, while CitiLink-Minutes tests model performance on realistic, less canonical discourse. Evaluating across both allows us to assess robustness and generalization of segmentation models across different domains, languages, and discourse structures.

\paragraph{Baselines}
We compare our NSP-based approach against several representative baselines. TextTiling~\cite{hearst1997texttiling} is a foundational unsupervised method that segments text by detecting lexical cohesion shifts between adjacent blocks, that has long served as a reference for discourse-level segmentation. Att+CNN \cite{badjatiya2018attention} is a supervised neural model that combines convolutional and bidirectional LSTM layers with attention to predict sentence-level segment boundaries. TopSeg \cite{ghinassi2023topicsegmentation} is a coherence-driven model trained with contrastive learning, optimizing sentence embeddings so that sentences within the same segment are closer in representation space, while sentences across segments are pushed apart. Topic boundaries are then identified where the cosine similarity between adjacent sentences falls below a threshold.

We also explored an alternative LLM-based segmentation paradigm inspired by \textit{LumberChunker} \cite{duarte2024lumberchunker}, using Gemini 2.5 Flash Lite \cite{comanici2025gemini25pushingfrontier} as the underlying generative model. This allows us to evaluate our approach in a cross-paradigm setup and assess its robustness against generative LLM-based segmentation. Other generative LLM-based methods (e.g., \citet{malkiel2024segllm}) are not included, as they typically generate free-form text rather than producing explicit boundary offsets, preventing evaluation with boundary-based metrics. For all included baselines, we use publicly available implementations and recommended hyperparameters whenever possible to ensure fair and reproducible comparisons.

\paragraph{Metrics}
We evaluate segmentation quality using several complementary metrics.  $P_k$ \cite{beeferman1997} and \textit{WindowDiff} (WD) \cite{pevznerhearst2002critique} quantify boundary errors within a sliding window. $P_k$ measures incorrect same-segment assignments, whereas WD is less punitive for near-miss boundaries. To mitigate the harsh treatment of near-miss boundaries, we additionally report the boundary F$_{\beta}$-measure (B-F$_1$) and Boundary Similarity ($B$) \cite{fournier2013evaluating}. B-F$_1$ treats segmentation as sequence labeling, granting partial credit for approximate matches, while $B$ compares boundary sets directly, rewarding close alignments that $P_k$ and WD may overlook.

\section{Experiments and Results}
\label{sec:experiments_and_results}

In this section, we present a series of experiments designed to evaluate the performance of our proposed segmentation approach, \textbf{SegNSP}, under different settings. We first assess in-domain performance on standard benchmarks to establish overall effectiveness. Subsequently, we examine cross-domain generalization using leave-one-municipality-out validation on the CitiLink-Minutes corpus. Finally, we provide a detailed error analysis to investigate linguistic and positional factors influencing segmentation accuracy.

\subsection{Training Setup}
\label{sec:experimental_setup}

We conducted experiments on two datasets to evaluate both in-domain and cross-domain performance. For WikiSection, we used the predefined splits of the \_en\_city subset. For CitiLink-Minutes, we followed the chronological split of the corpus, assigning 60\%, 20\%, and 20\% of documents for training, validation, and testing, respectively, reserving the most recent meeting minutes for testing to simulate realistic deployment conditions.

SegNSP is initialized from a pre-trained BERT encoder\footnote{https://huggingface.co/neuralmind/bert-base-portuguese-cased} and fine-tuned for sentence continuity prediction using the proposed segmentation-aware objective described in Section~\ref{sec:our_approach}. Hyperparameters, including learning rate, batch size, and loss weighting coefficients, were optimized on the validation split, resulting in a learning rate of $5\times10^{-6}$, batch size of $8$, focal parameters $\gamma{=}1.5$ and $\alpha{=}0.8$, confidence penalty of $0.15$, and a boundary weight of $0.2$, with early stopping applied after $12$ epochs. Boundary detection during inference uses a threshold $\tau{=}0.5$, selected based on validation B-F$_1$ performance.

Training follows the intra/inter-segment ratios and harder negative sampling strategy described in Section~\ref{sec:our_approach}, ensuring the model effectively captures both semantic continuity within segments and transitions across topics. All baselines were trained or tuned using their recommended configurations and publicly available implementations to ensure fair and reproducible comparisons.

\subsection{Results} \label{sec:results}

We evaluated cross-dataset performance by comparing our approach against different baselines on CitiLink-Minutes and WikiSection\_en\_city (Table~\ref{tab:segmentation-results}). On CitiLink-Minutes, SegNSP achieved the highest B-F$_1$ (0.79) and Boundary Similarity $B$ (0.59), indicating strong alignment between predicted and annotated topic transitions. Complementary error-based metrics reinforce this finding, with $P_k{=}0.08$ and WindowDiff (WD){=}0.10.
On WikiSection\_en\_city, SegNSP also attained the highest B-F$_1$ (0.65) while maintaining competitive $P_k$ (0.14) and WD (0.18), demonstrating that modeling sentence-continuity generalizes beyond the administrative domain. Notably, the multi-task model by \citet{lee2023} (combining NSP, same-topic prediction, and topic classification) reported the lowest error metrics on WikiSection, but it was excluded from our B-F$_1$ comparison due to its reliance on auxiliary topic labels and a lack of publicly available code for replication.

In contrast, the LLM-based LumberChunker performed substantially worse on CitiLink-Minutes and did not improve over the strongest encoder-based baselines on WikiSection\_en\_city, highlighting practical challenges of generative segmentation approaches for tasks requiring precise boundary detection (Section~\ref{sec:related_work}). Overall, these results support NSP as a lightweight, encoder-based alternative that balances high boundary accuracy with computational efficiency across domains.

\begin{table*}[t]
\centering
\caption{Segmentation results across datasets. Higher B-F$_1$/$B$ and lower $P_k$/WD indicate better performance. Values are macro-averaged; best in bold.}
\label{tab:segmentation-results}
\begin{tabular}{l *{8}{c}}
\toprule
\multirow{2}{*}{\textbf{Model}} &
\multicolumn{4}{c}{\textbf{CitiLink-Minutes}} &
\multicolumn{4}{c}{\textbf{WikiSection\_en\_city}} \\
\cmidrule(lr){2-5}\cmidrule(lr){6-9}
& \textbf{B-F$_1$} $\uparrow$
& \textbf{B} $\uparrow$
& \boldmath{$P_k$} $\downarrow$
& \textbf{WD} $\downarrow$
& \textbf{B-F$_1$} $\uparrow$
& \textbf{B} $\uparrow$
& \boldmath{$P_k$} $\downarrow$
& \textbf{WD} $\downarrow$ \\
\midrule
TextTiling        & 0.15 & 0.08 & 0.39 & 0.44 & 0.09 & 0.05 & 0.44 & 0.46 \\
Att+CNN           & 0.34 & 0.21 & 0.25 & 0.27 & 0.14 & 0.08 & 0.24 & 0.27 \\
TopSeg            & 0.42 & 0.26 & 0.18 & 0.25 & 0.48 & 0.34 & 0.15 & 0.19 \\
LumberChunker     & 0.10 & 0.06 & 0.34 & 0.35 & 0.42 & 0.29 & 0.31 & 0.35 \\
STP + TC + NSP    & --   & --   & --   & --   
                  & --   & --   & \textbf{0.05} & \textbf{0.05} \\
\textbf{SegNSP}               & \textbf{0.79} & \textbf{0.59} & \textbf{0.08} & \textbf{0.10}
                  & \textbf{0.65} & \textbf{0.47} & 0.14 & 0.18 \\
\bottomrule
\end{tabular}
\end{table*}

We further evaluated SegNSP in an in-domain setting using the CitiLink-Minutes dataset to assess cross-municipality generalization. Specifically, we employed a leave-one-municipality-out cross-validation across the six municipalities (Alandroal, Campo Maior, Covilhã, Fundão, Guimarães and Porto). In each fold, five municipalities were used for training and the remaining one for evaluation, ensuring that the model was tested on an unseen domain. This setup enables the assessment of SegNSP's ability to generalize across stylistical, structural, and topical variations. 

Table~\ref{tab:leave-one-out-results} shows noticeable variation in performance across municipalities. While SegNSP achieved strong results on several cases, performance dropped for others, reflecting the challenges of generalizing to domains with distinct local conventions. These findings suggest that, although the model effectively captures general segmentation patterns, additional domain adaptation may be necessary to achieve consistent performance across all municipalities.

\begin{table*}[t]
\centering
\caption{Leave-one-municipality-out cross-validation results on CitiLink-Minutes. Higher B-F$_1$/B and lower $P_k$/WD indicate better performance.}
\label{tab:leave-one-out-results}
\begin{tabular}{@{}l @{\hspace{1em}} *{6}{N}@{}}
\toprule
\textbf{Measure} & \textbf{Alandroal} & \textbf{Campo Maior} & \textbf{Covilhã} & \textbf{Fundão} & \textbf{Guimarães} & \textbf{Porto} \\
\midrule
B-F$_1$ $\uparrow$ / B $\uparrow$ &
{0.74 / 0.49} & {0.77 / 0.56} & {0.72 / 0.50} & {0.24 / 0.15} & {0.68 / 0.48} & {0.34 / 0.19} \\
$P_k$ $\downarrow$ / WD $\downarrow$ &
{0.08 / 0.09} & {0.09 / 0.13} & {0.10 / 0.12} & {0.15 / 0.21} & {0.07 / 0.12} & {0.11 / 0.20} \\
\bottomrule
\end{tabular}
\end{table*}

\subsection{Error Analysis}
To move beyond aggregate performance metrics, we conducted a qualitative error analysis across both datasets. While WikiSection served as a controlled, established benchmark, CitiLink-Minutes reflected the structural variability and administrative boilerplate typical of real-world local governance records.

We first examined the distribution of $P(\texttt{not\_next})$ across all consecutive sentence pairs. On CitiLink-Minutes (see Figure~\ref{fig:prob_citilink} in \ref{sec:appendix}), the model exhibited a sharply bimodal distribution, with only 1.4\% of predictions falling within the ambiguous interval $[0.4, 0.6]$. In contrast, WikiSection (see Figure~\ref{fig:prob_wiki} in \ref{sec:appendix}) showed a broader overlap ($4.7\%$) and a negative separation gap ($-0.015$), reflecting greater uncertainty in boundary decisions. This pattern suggests that semantic transitions in  Wikipedia articles are often more gradual and less structurally explicit than in administrative meeting minutes.

A more granular analysis of positional error distributions (see Figure~\ref{fig:combined_pos} in \ref{sec:appendix}) revealed further dataset-specific trends. In CitiLink-Minutes (see Figure~\ref{fig:error_citilink} in \ref{sec:appendix}), over-segmentation (false positives) increased toward the final 10\% of documents, corresponding largely to administrative boilerplate, such as adjournment formalities, that is misclassified as topical transitions. Conversely, WikiSection exhibited a predominance of under-segmentation errors (false negatives) near document endings, as shown in Figure~\ref{fig:error_wiki}, consistent with shorter sections and more gradual topic transitions.

Examining errors at the municipality level (Figure~\ref{fig:over_under}) highlighted local stylistic effects within the CitiLink-Minutes. Alandroal showed a higher rate of under-segmentation (FN $\approx 0.065$), indicating a conservative boundary detection that may miss subtle topic shifts. Campo Maior, by contrast, exhibited higher over-segmentation rates (FP $> 0.05$), potentially influenced by repetitive linguistic patterns. Fundão and Porto displayed near-zero error rates, indicating that the chosen decision threshold $\tau$ generalized effectively to these municipalities.

Overall, these analyses indicate that segmentation performance is shaped not only by domain characteristics but also by localized stylistic conventions and structural variability. These findings motivate future work on adaptive thresholding or municipality-aware calibration to improve robustness across heterogeneous administrative records.

\section{Conclusions \& Future Work}
\label{sec:conclusions}

In this paper, we revisited the classic next sentence prediction (NSP) objective for linear text segmentation, showing that by redefining NSP with a segmentation-aware loss and domain-specific fine-tuning yields a lightweight but effective approach for identifying topical boundaries. Our results demonstrate that this label-agnostic model is highly competitive across domains, achieving a B-$F_1$ of 0.65 on WikiSection and outperforming established, reproducible baselines like TopSeg and Att+CNN by a significant margin. Moreover, our evaluation on the CitiLink-Minutes corpus highlights the applicability of this approach to real-world deliberative records, where discourse structure is long-range, heterogeneous, and loosely standardized.

Our error analysis further reveals that, despite its strengths, NSP-based segmentation is inherently locally focused and lacks explicit modeling of global discourse structure. While the model exhibits a strong bimodal probability distribution, it also displays positional bias, particularly a tendency toward over-segmentation near document conclusions (90--100\% range), often associated with administrative boilerplate. Future work will explore hybrid architectures that extend NSP's local coherence modeling with hierarchical smoothing or attention mechanisms to better capture gradual topic shifts. Finally, we plan to evaluate NSP-based segmentation as a specialized chunking strategy for retrieval-augmented generation. We hypothesize that high-precision boundary detection will yield more semantically coherent passages, thereby improving embedding quality for dense retrieval and reducing contextual noise in downstream generation tasks.

\section{Limitations}
\label{sec:limitations}
While our NSP-based approach demonstrates strong performance across both administrative and expository datasets, it exhibits several limitations that warrant further investigation.

First, our model is primarily locally focused. By operating on consecutive sentence pairs $(s_i, s_{i+1})$, the classifier lacks an explicit representation of the broader, global discourse structure. This can lead to locally constrained boundary decisions, where long-range thematic dependencies spanning multiple paragraphs are missed.

Second, a distinct positional bias was observed in our error analysis. Specifically, in the CitiLink-Minutes dataset, the model exhibits a spike in over-segmentation within the final 10\% of documents. This is largely caused by administrative boilerplate and adjournment protocols, which share linguistic markers with topical transitions, leading the model to trigger false boundaries.

Third, although our model is label-agnostic and does not require topic annotations, it still depends on high-quality sentence tokenization. In domains with highly irregular punctuation or OCR errors, the formation of the $(s_i, s_{i+1})$ pairs may be compromised, directly affecting the classifier's accuracy.

Fourth, our cross-dataset evaluation on WikiSection was restricted to the \texttt{en\_city} subset. While this allowed for a focused comparison against administrative records, it leaves the model’s performance on other domains (e.g., \texttt{en\_disease}) and other languages (e.g., German subsets \texttt{de\_city}, \texttt{de\_disease}) unexplored. Future work should investigate whether the observed bimodal distribution and semantic density challenges remain consistent across these diverse topical and linguistic categories.

Finally, while we establish strong benchmarks on CitiLink-Minutes ($B\text{-}F_1$ = 0.79) and WikiSection ($B\text{-}F_1$ = 0.65), our comparison with the multi-task model by \citet{lee2023} is limited by the lack of publicly available code for that framework. Consequently, we could not evaluate their model on the Portuguese corpus or perform a direct comparison using the $B\text{-}F_1$ metric on WikiSection. Furthermore, our evaluation of generative paradigms was limited to Gemini 2.5 Flash Lite. As such, we did not explore more recent or larger-scale generative models like Gemini 3 or GPT-5. A broader comparison with these state-of-the-art LLMs could provide further insights into the trade-offs between zero-shot generative reasoning and our specialized, encoder-based approach.

\section{Ethical Considerations}
\label{sec:ethical}
\paragraph{Intended Use and Misuse Potential} The SegNSP model was developed to advance research on linear text segmentation, specifically within the context of deliberative and administrative discourse. Our work aims to support researchers and civic organizations in improving the accessibility of municipal records through automated structure induction, thereby fostering government transparency. While our focus is on structural coherence, misuse risks exist, such as the automated reconstruction of sensitive or redacted discussions if applied to non-public documents. We encourage users to apply these methods only to documents where public disclosure is intended.

\paragraph{Usage of AI} AI tools were used to assist in the writing process, specifically to enhance clarity and readability, and to improve language flow and presentation.

\paragraph{Environmental Impact} The computational demands of modern NLP models contribute to significant energy consumption and CO$_2$ emissions. To minimize this impact, we focused on a lightweight, encoder-based approach (SegNSP) rather than massive generative models for our primary architecture. While our LLM-based baseline (LumberChunker) utilized the Gemini 2.5 Flash Lite model in a zero-shot setting, the majority of our experiments involved fine-tuning a BERT-base encoder. These experiments were conducted on local GPU infrastructure, incurring approximately 100 GPU hours, including hyperparameter tuning, cross-validation folds on CitiLink-Minutes, and final evaluations on WikiSection. Furthermore, LLM-based experimentation was performed using the Google Cloud Platform, requiring approximately 14 hours of inference time.

\section*{Acknowledgments}
This work was funded within the scope of the project  CitiLink, with reference 2024.07509.IACDC, which is co-funded by Component 5 - Capitalization and Business Innovation, integrated in the Resilience Dimension of the Recovery and Resilience Plan within the scope of the Recovery and Resilience Mechanism (MRR) of the European Union (EU), framed in the Next Generation EU, for the period 2021 - 2026, measure RE-C05-i08.M04 - ``To support the launch of a programme of R\&D projects geared towards the development and implementation of advanced cybersecurity, artificial intelligence and data science systems in public administration, as well as a scientific training programme,'' as part of the funding contract signed between the Recovering Portugal Mission Structure (EMRP) and the FCT - Fundação para a Ciência e a Tecnologia, I.P. (Portuguese Foundation for Science and Technology), as intermediary beneficiary.\footnote{\url{https://doi.org/10.54499/2024.07509.IACDC}}

\bibliography{custom}

\appendix

\section{Appendix A}
\label{sec:appendix}

\begin{figure*}[t]
    \centering    \includegraphics[width=0.9\textwidth]{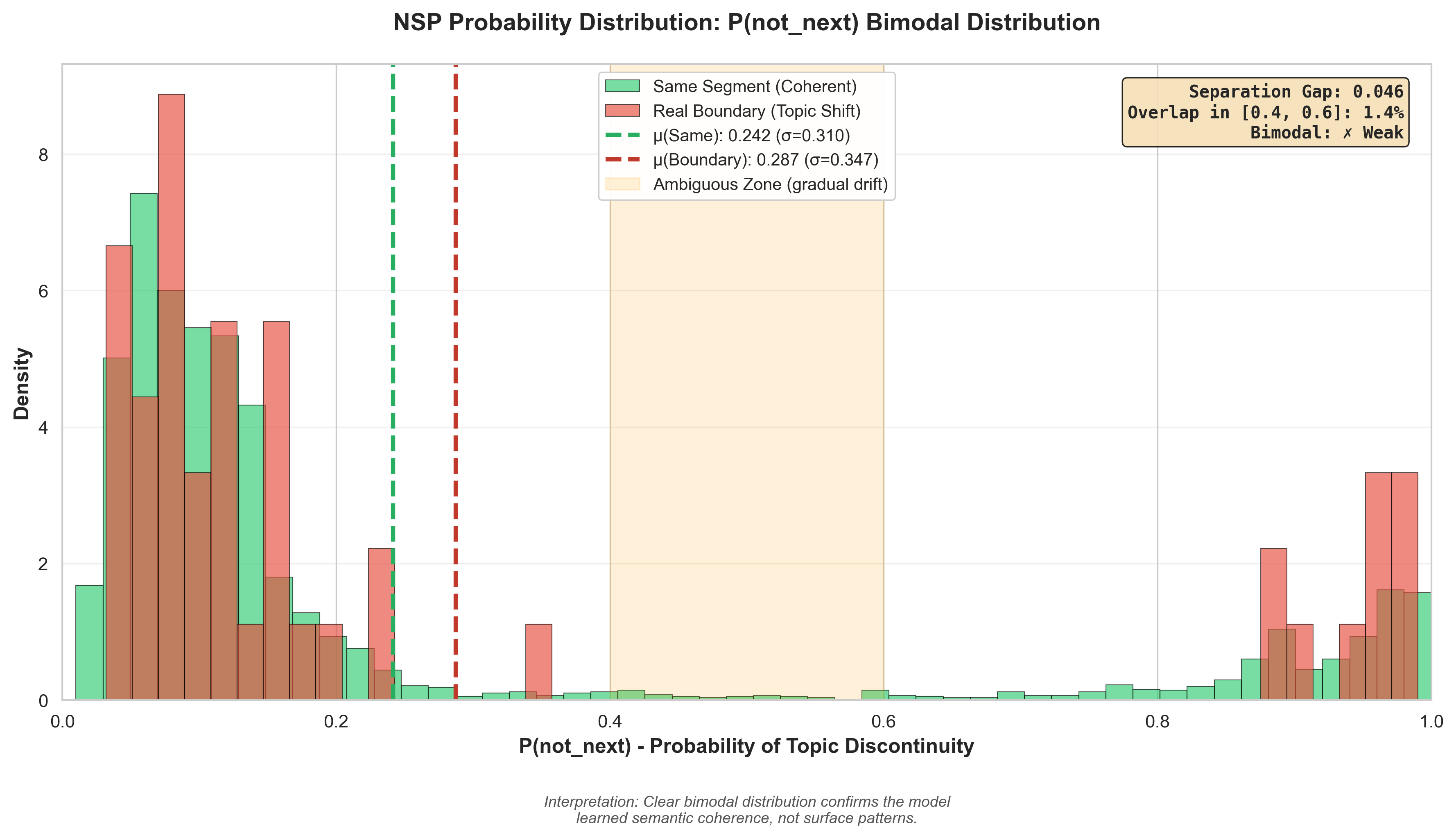}
    \caption{CitiLink-Minutes: bimodal distribution of $P(\text{not\_next})$, showing a 1.4\% overlap in the ambiguous zone $[0.4, 0.6]$.}
    \label{fig:prob_citilink}
\end{figure*}

\begin{figure*}[h]
    \centering    \includegraphics[width=0.9\linewidth]{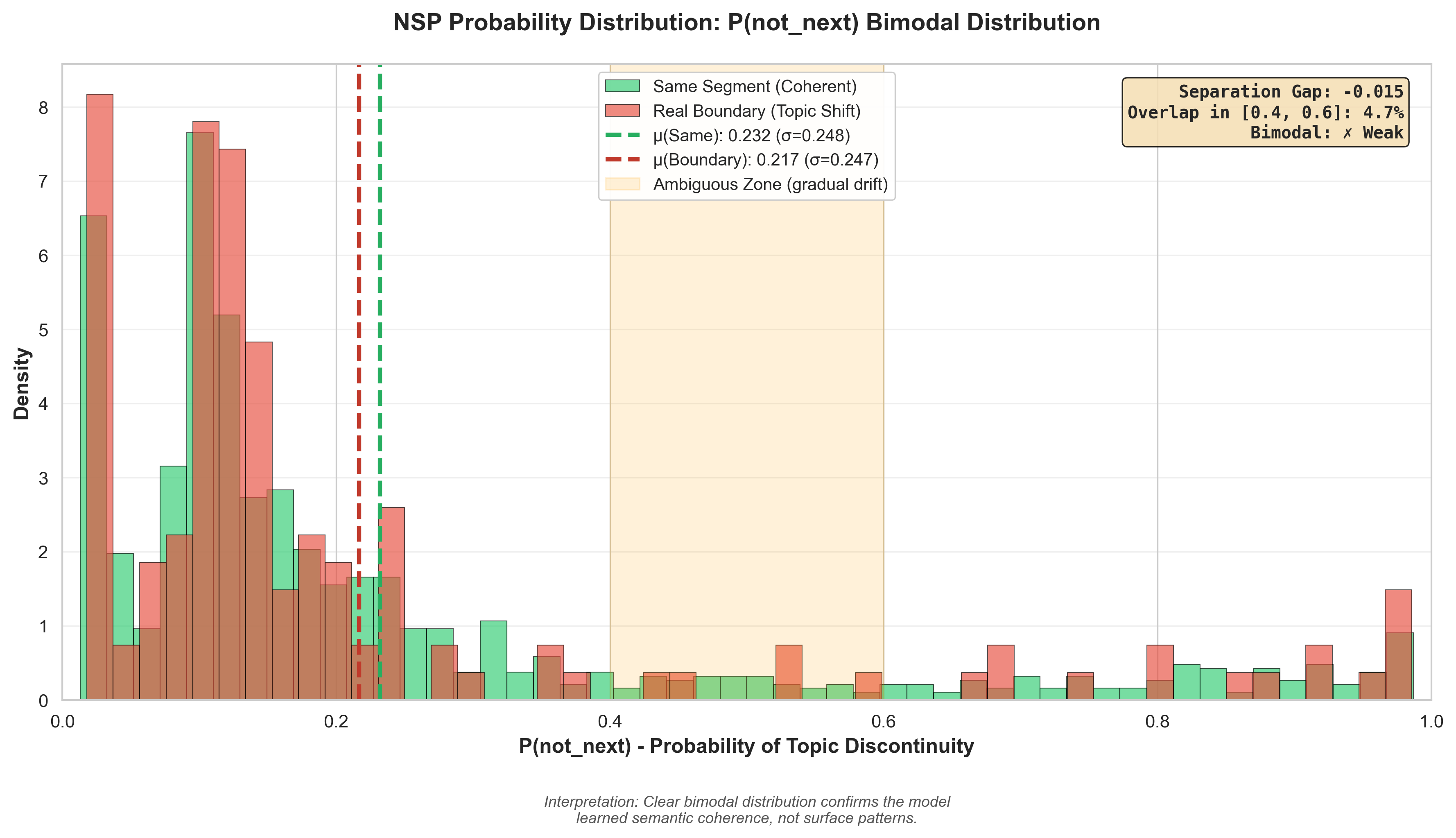}
    \caption{WikiSection: bimodal distribution of $P(\text{not\_next})$, broader overlap (4.7\%) in the ambiguous zone $[0.4, 0.6]$..}
    \label{fig:prob_wiki}
\end{figure*}


\begin{figure*}[!b] 
    \centering
    \begin{subfigure}[b]{0.6\textwidth}
        \centering
        \includegraphics[width=\linewidth]{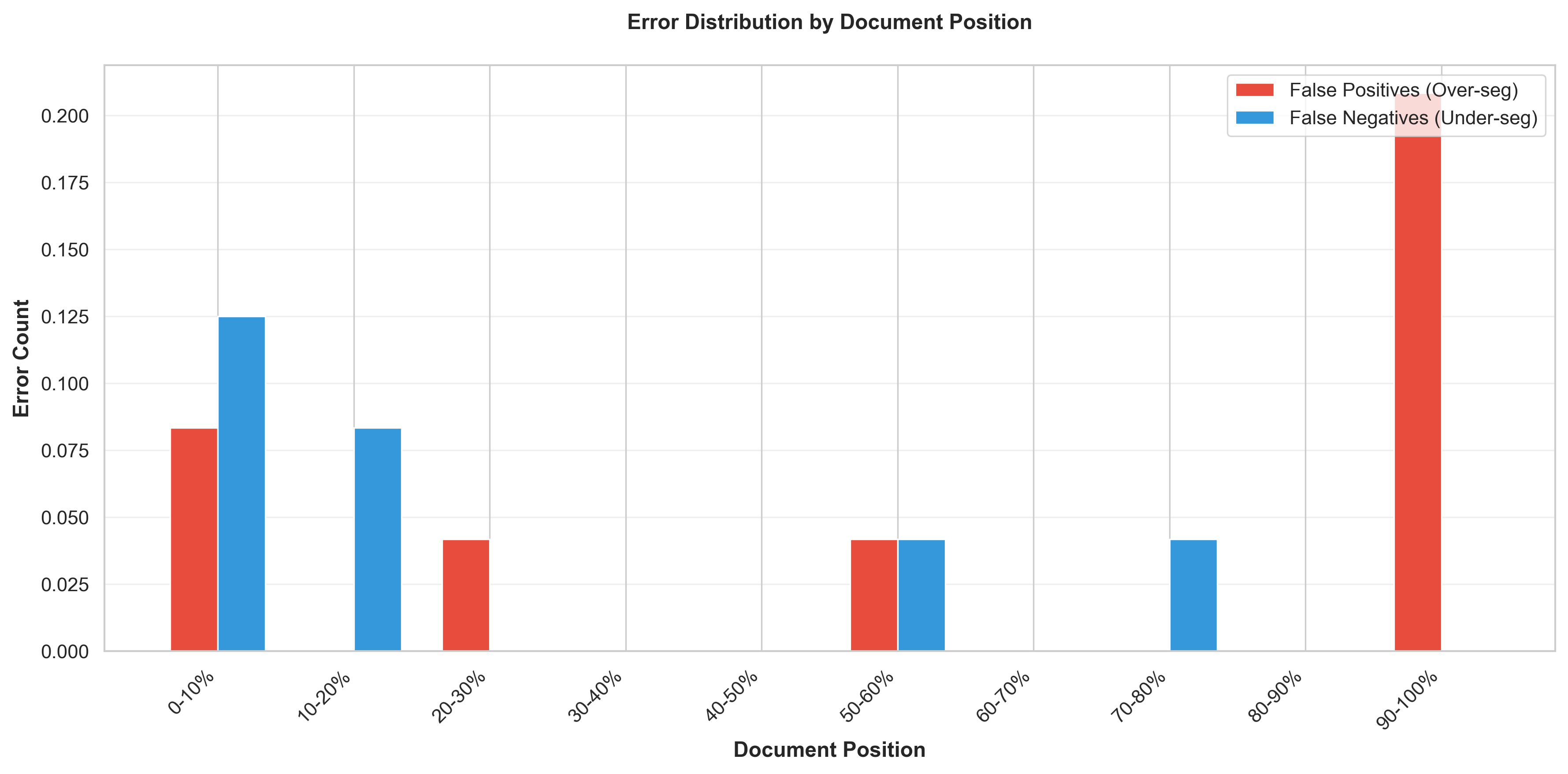}
        \caption{CitiLink-Minutes: error distribution across document positions, highlighting the 90--100\% over-segmentation spike.}
        \label{fig:error_citilink}
    \end{subfigure} \hfill

    \begin{subfigure}[b]{0.6\textwidth}
        \centering
        \includegraphics[width=\linewidth]{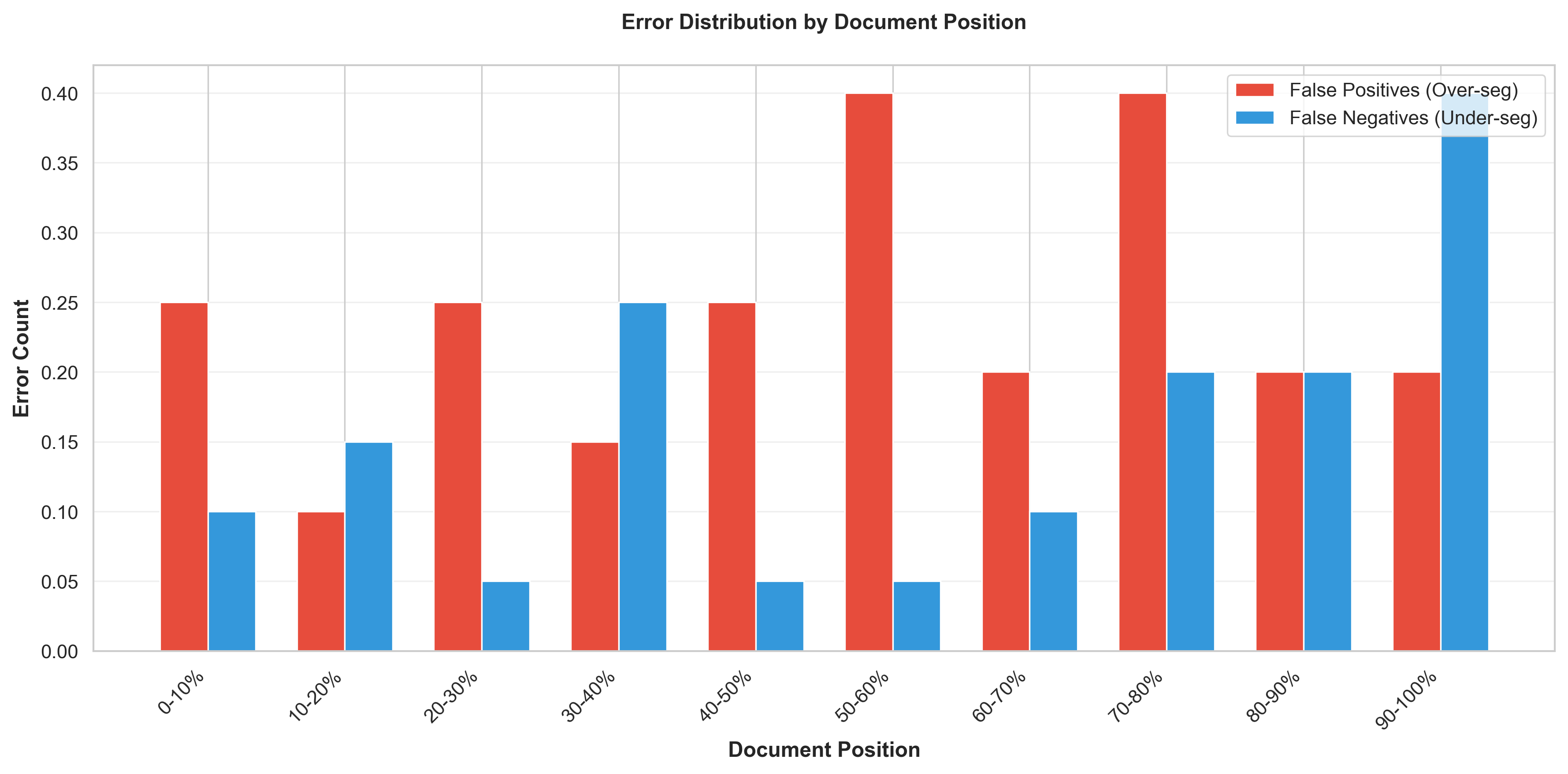}
        \caption{WikiSection: error distribution across document positions.}
        \label{fig:error_wiki}
    \end{subfigure}
    \caption{Distribution of segmentation errors by document position.}
    \label{fig:combined_pos}
    \begin{minipage}{0.65\textwidth} 
        \centering
        \includegraphics[width=\linewidth]{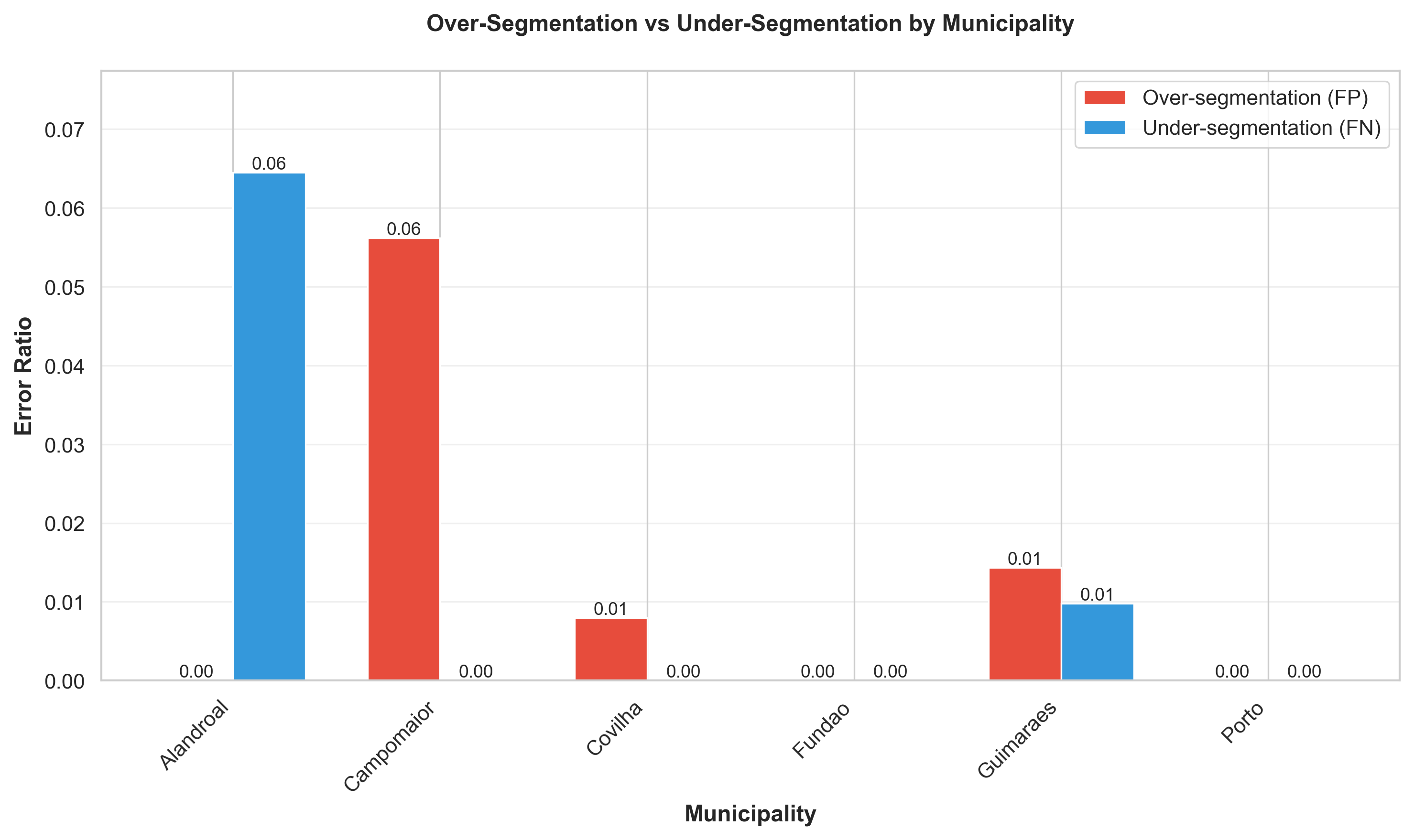}
        \caption{Over- and under-segmentation ratios across CitiLink-Minutes municipalities.}
        \label{fig:over_under}
    \end{minipage}
\end{figure*}

\end{document}